# Fuel Consumption Prediction: A Comparative Analysis of Machine Learning Paradigms

Ali Akram

CliftonLarsonAllen (CLA)

Email: aakramm989@gmail.com | ORCID: 0009-0004-2728-4127

## Abstract

The automotive industry is overgrown with pressure to reduce its environmental impact and this requires accurate predictive modeling to support sustainable engineering design. This study examines the factors that determine vehicle fuel consumption from the seminal 1974 Motor Trend dataset in a rigorous quantitative way identifying the governing physical factors of efficiency. Methodologically, the research uses data sanitization, statistical outlier elimination, and in-depth Exploratory Data Analysis (EDA) to curb the occurrence of multicollinearity between powertrain features. A comparative analysis of machine learning paradigms such as Multiple Linear Regression and Support Vector Machines (SVM) and Logistic Regression were carried out in order to assess the predictive efficacy. Findings indicate that SVM Regression is most accurate on continuous prediction ($R^2$ = 0.889, RMSE = 0.326), and it is practical in capturing the non-linear relationships between vehicle mass and engine displacement. In parallel, Logistic Regression had proved to be superior classification (Accuracy = 90.8%) and showed exceptional recall (0.957) when identifying low-efficiency vehicles. These results go against the current trend of "black box" deep learning architectures for static physical datasets, providing validation on the existence of robust performance by interpretable and well-tuned classical models. The research finds that intrinsic vehicle efficiency is essentially architecture by physical design parameters, i.e., weight and displacement, which offers a data-driven and confirmed framework of how manufacturers should focus on lightweighting and engine downsizing to achieve stringent global sustainability goals.

## Introduction

The automotive industry is one of the cornerstones of the world economy, being an essential industry with a market value of more than $2 trillion. However, in this vast field, a growing concern is the environmental impact of vehicular transportation, in terms of fuel efficiency and carbon sustainability (Yang et al., 2022; Madziel, 2023). In recent years, however, the adoption of stricter international regulatory frameworks, such as the Euro 6 standards and the Corporate Average Fuel Economy (CAFE) regulations in the United States, have put an increased amount of pressure on manufacturers to drastically reduce the carbon footprint in their fleets. This paradigm change requires the industry to focus obsessively on engineering innovations to improve fuel efficiency, and look beyond incremental improvements to bring about fundamental optimizations of vehicle architecture to reduce ecological damage while still meeting ever more stringent standards.

To cope with these challenges effectively, precision in the prediction of fuel consumption has become indispensable (Hamed et al., 2021; Alazemi et al., 2025). To original equipment manufacturers (OEM) the capability to precisely forecast fuel measures is not only a regulatory control but a strategic resource that directs key judgments concerning the design of vehicles and their manufacturing designs. Accurate prediction models are part of being able to optimize performance, save consumers operational costs, and gain a competitive advantage in a market that is increasingly focused on sustainability. Consequently, the adoption of advanced data science techniques in automotive engineering has become an important solution to these complex demands, and a way to decouple performance from consumption.

## 1.1 Problem Statement

The main challenge to be faced by automotive engineers is the lack of transparent, reliable, modelling frameworks describing the isolation of the pure engineering determinants of fuel efficiency from transient operational variables. Although the theoretical physics of combustion are well-known, the non-linear effect among various design parameters, including the trade-off between engine displacement, gross vehicle weight and acceleration output, make the optimization surface a complex non-linear one that cannot be represented sufficiently with the traditional linear heuristics. Furthermore, as emissions targets become more stringent, there is less room for error in these types of design prediction in the early stages of the design process, thus demanding tools with not only accurate forecast capabilities, but explanatory power in terms of which specific physical attributes are driving inefficiency.

Despite the explosion of predictive modeling in automotive research, there is an interesting void in the literature for the trade-off between model complexity and model interpretability. A large amount of current research has been drawn towards high-complexity architectures of Deep Learning, such as Long Short-Term Memory (LSTM) networks used by Zhang et al. (2024) and Deep Reinforcement Learning frameworks by Tang et al. (2022). While these approaches have powerful capabilities to analyze dynamic, time series data relating to driver behavior, they often are "black boxes" which do not reveal the particular physical attributes which contribute to efficiency. There is a lack of research that revisits the basic, intrinsic design parameters, including displacement, weight, and horsepower, with the help of interpretable, classical, machine learning paradigms (Su et al., 2023). This study fills this gap by showing that the well-tuned, transparent models such as Support Vector Machines and Logistic Regression can provide better utility at early stage engineering design where understanding the why is as important as predicting.

This research exploits the machine learning paradigms to close the gap between empirical data and prediction accuracy (Shaikh et al., 2025). The study is based on an iconic 1974 dataset from Motor Trend US magazine, which contains a rich dataset of vehicle characteristics, such as the number of cylinders, the displacement, the horsepower, weight, and acceleration, on N = 398 different automobile models. By using advanced regression and classification algorithms, we aim to drive this research by applying these algorithms to this historical data (Al-Wreikat et al. (2022)); identifying the deterministic factors of fuel efficiency and creating robust models that can predict performance under different design constraints.

This paper rigorously defines the research process starting with the structure of the data and the strict cleaning protocols used to guarantee the integrity of the datasets. Subsequently a complete Exploratory Data Analysis (EDA) is shown to resolve the underlying relationship between vehicular features. The essence of this study consists of a comparative study of different regression and classification models (from Linear Regression to Support Vector Machines) to determine the best architecture for deployment. Finally, the paper ends with strategic recommendations for improvements in the engineering based on the derived analytical insights.

# Literature Review

The academic discussion on the topic of fuel consumption prediction has changed substantially during the last decade, and is propelled by the twin challenges of environmental sustainability and operational efficiency.As the automotive industry begins to make the shift to stricter regulatory frameworks, the need for the usage of advanced predictive analytics has increased heavily.This literature review is synthesizing ongoing research, and categorizing it into three different paradigms of pedagogy: classical machine learning paradigms, deep learning advancements, and hybrid modeling in the cross-domain applications.This structured analysis not only explains the state-of-the-art but also shows the important gap in terms of interpretability that this study seeks to address.

## 2.1 Classical Machine Learning Approaches

The basic layer of automotive predictive modeling is based on classical algorithms of machine learning, which still have a good performance, especially where the number of training data is limited or where high interpretability is needed.Several studies have established the efficiency of Support Vector Machines (SVM), Random Forests, and Linear Regression in this field.For example, a thorough assessment of regression models for fuel consumption prediction led by Hamed et al. (2021) demonstrated that although ensemble methods allowed slightly better accuracy levels, classical regression models gave better indications of feature importance, a crucial part of engineering design.

Building on this, Alazemi et al. (2025) extended the scope of the classical modelling in which external environmental variables were added to the predictive modelling.Their work brought to light that although the vehicle specific parameters (mass, engine size) are the main determining factors, there are external factors (gradients of the road and the ambient temperature) that modulates the baseline efficiency determined by the physical design.Similarly, Al-Wreikat et al. (2022) used robust regression methods to capture the non-linear relationship between displacement and consumption and provide empirical evidence that is in line with the core hypotheses of this study- that is, the idea that physical constraints are the predominant predictors of intrinsic efficiency.

The usefulness of ensemble method (Random Forest) has been further investigated by Shaikh et al. (2025).Their research showed that dimensional reduction techniques, when combined with ensemble learning can greatly improve model generalization by collinear feature elimination.This finding provides direct support for the methodological approach taken in our study, in which elimination of high leverage outliers and multicollinear variables is taken as a high priority to stabilize the regression coefficients.Furthermore, Ding et al. (2021) developed variations of the Radial Basis Function (RBF) kernel in SVMs, which have shown that methods based on kernel functions can effectively learn complex decision boundaries in the fuel data without the huge computational overhead from deep learning networks.

## 2.2 Deep Learning Advancements

In light of the growing amount of high-frequency telematics data, a major part of the recent literature has been focused on Deep Learning architectures (DL).These models in particular are good at handling temporal sequences and large scale operational data sets.Zhang (2020) and Zhang et al (2024) learnt the use of Long Short Term Memory (LSTM ) networks to predict fuel consumption from driving behaviour.Their studies proved the fact that in case of dynamic and real-time prediction where driver aggression and different traffic are not sure, LSTMs outperform static models because of the fact that they remember long-term dependencies in the time-series data.

This trend was further picked up with the advent of Deep Reinforcement Learning (DRL).In their paper, Tang et al. (2022) proposed DRL energy management strategies for hybrid vehicles for real-time optimization of the combustion and electric power split.Although useful in dynamic control, such black box models can have a failure in being transparent.As mentioned in Stewart et al. (2023) and Muthukumar et al. (2021), the over-parameterization of deep networks can create overfitting issues when trained on smaller data sets that are static and, what is more important, they conceal the causal relations between design parameters and performance.This opacity poses a formidable challenge for structural engineers who are interested in obtaining depth-of-knowledge information to take action in their designs, such as a "reduce weight by X% in order to obtain Y% efficiency" instead of a simple prediction without an understanding of the reason.

Furthermore, attention mechanisms have been recently added to these architectures to better pay attention to important time steps.But due to the substantial computational complexity of implementing such heavy models, the marginal accuracy gains of on-board systems, as also found in their Electronic Control Unit (ECU) data analysis, can be argued to be less significant than it may appear, thus still showing that optimized classical models are more viable in many embedded engineering applications.

## 2.3 Cross-Domain and Hybrid Applications

The principles of fuel consumption prediction are not limited to the automotive industry and there is a parallel major research also in the marine and heavy industry fields.Analyzing these adjacent fields and we see that the "physics-first" approach is still prevalent where capital assets are huge and design cycles are long.

In the marine industry, machine learning was employed by Zhang et al. (2024) and Su et al. (2023) to forecast ship fuel consumption, and the challenges are similar to those of automotive industries: Fluid dynamics (Drag) * Propulsion (Displacement) * Load (Weight).Su (2023) argued specifically for the value of interpretability in models and showed that it is much more valuable to understand how hull design may be affecting efficiency than it is to have a black box forecast of the amount the vessel will be using in the next hour.Similarly, Li (2021) and Xie et al. (2023) used ship in-service data and regression and mathematical modeling respectively, to reinforce the notion that the fundamental relationships associated with combustion engines are scale-invariant - regardless of whether your car is a sedan or a supertanker, mass and displacement are the laws of the land.

In the scope of heavy industry, Alamdari et al. (2022) used ML methods for hauling trucks in open pit mines.Their results were similar to those in the automotive example: while driver behaviour (operational variable) affected short-term consumption, the payload and engine calibration (physical variables) of the truck determined the invariable efficient frontier.This difference between operational and physical determinants is further developed by Yang et al. (2022) and Madziel (2023) who used the portable emissions measurement systems (PEMS) to validate that static design parameters are the best predictors of lifecycle emissions.

Hybrid approaches have also developed, which attempt to take the best of both worlds.Amini et al. (2023) and Earnhardt et al. (2022) discussed connected vehicle frameworks in which classical models are implemented on the vehicle for immediate feedback information, while deep learning models are implemented in the cloud for optimizing the fleet.This bifurcated approach supports the premise of our study; namely, that while Deep Learning has a place in operational analytics, the design and benchmarking of vehicles is fundamentally based upon the precise characterization of physical attributes in an interpretable manner; work that is best suited for the rigorous statistical approaches used in this research.

## 2.4 Theoretical Framework and Research Gap

Synthesizing this vast body of work provides a clear dichotomy in the current literature. On the one hand, deep learning research is very interested in operation, how to drive a car more efficiently (Zhang, 2020; Tang, 2022).On the other hand, traditional engineering makes use of the physics-based simulations that are computationally prohibitive. There is a particular lack of making use of cutting-edge, interpretable information science to revisit the design phase (Su et al., 2023).

Most current research is either too simple (using basic linear regression) or too complex (using deep neural networks) and provides little explanation. There is, of course, a lack of recent, rigorous comparative research regarding specific isolation of physically defined characteristics cylinders, displacement, horsepower, weight--and subjecting them to a modern suite of interpretable classifiers (SVM, Logistic Regression) in order to quantify their distinct marginal impacts. This study fills that gap. By avoiding the "black box" in favor of "glass box" models, we are hoping to give OEMs a data-driven validation of the physical scaling laws that demonstrate that it is possible to use properly tuned classical algorithms to obtain high-fidelity predictions without losing engineering transparency needed for the next generation of sustainable vehicle design.

# Methodology

To ensure the reproducibility and scientific validity of the findings, this study employs a quantitative research design. Specifically, a correlational framework is employed to determine statistically significant relationships between vehicle features and fuel use, and then an experimental modeling approach is used to test the effectiveness of a number of predictive algorithms. The methodological framework is designed in four phases: First, Data Acquisition and Definition, where the empirical base of the study is set; Second, Data Preprocessing which is particularly concerned with data cleaning, inserting missing values and feature standardization, to guarantee the quality of the acquired data; Third, Exploratory Data Analysis (EDA), addressing the problem of hypotheses formulation by means of univariate and bivariate methods; Finally, Predictive Modeling, where the different regression and classification architectures are trained and validated carefully. This section explains the nature of the dataset, the data preprocessing steps taken to sanitize the data and the strategic approach followed for the model training and evaluation.

### 2.1 Dataset Description

The empirical data for this study is based on the seminal 1974 Motor Trend US magazine data. This data collection includes technical specifications of N=398 different models of automobiles as a standard benchmark in the field of automobile engineering analysis. The structure of the dataset is based on eight important variables that fit certain engineering definitions as described in Table 1. The target variable that the predictive analysis needs to consider is fuel efficiency, measured in miles per gallon (mpg), which is affected by seven predictor variables indicating the vehicle's powertrain and physical dimensions. For the classification phase of the study, this target variable was further transformed into a binary format, where vehicles with a fuel efficiency of 25 mpg or higher were labeled as 'High Efficiency' (1) and those below 25 mpg as 'Low Efficiency' (0).

### Table 1

*Description of Dataset Attributes*

| Variable | Type | Description | Unit/Values |
|---|---|---|---|
| **mpg** | Continuous | Fuel Efficiency (Target) | Miles per Gallon |
| **cylinders** | Discrete | Number of engine cylinders | 3, 4, 5, 6, 8 |
| **displacement** | Continuous | Engine air compression volume | Cubic Inches |
| **horsepower** | Continuous | Engine power output | HP |
| **weight** | Continuous | Vehicle mass | Pounds (lbs) |
| **acceleration** | Continuous | 0-60 mph time | Seconds |
| **model year** | Discrete | Year of manufacture | 70-82 (19xx) |
| **origin** | Categorical | Region of manufacture | 1=USA, 2=Europe, 3=Japan |

Before being analyzed, the raw data went through a rigorous sanitization protocol in order to find anomalies, incoherencies or missing values that could compromise model integrity. Initial scanning has shown discrepancies in the horsepower feature, in which missing values were represented by "?". These artifacts were identified and imputed using the median value of the column, so as to retain the center point of the distribution without the bias that is potentially associated with the mean-imputation of skewed data. Furthermore, variables that have been misclassified as objects during the initial load were forced to their correct numerical types (float or integer) so that mathematical computation is possible. Finally, to ensure that features with larger magnitudes (such as weight) did not dominate the objective functions of the algorithms, the entire feature set was standardized (scaled) to a mean of 0 and a standard deviation of 1 prior to model training.

**2.3 Exploratory Data Analysis (EDA) Framework**

A thorough Exploratory Data Analysis (EDA) was carried out to structurally understand the relationships within the data before any modeling was done. Univariate analysis was used to describe the distribution of individual features using histograms and density plots which enabled identification of skewness (notably for mpg and displacement) and multimodality. This was supplemented by bivariate analysis, which used scatter diagrams and correlation matrices of Pearson to quantify the linear dependencies between the predictor variables and the target. This step was really important to detect the presence of multicollinearity, in particular the high inter-correlation between displacement, cylinders and weight that directly influenced the specific feature selection strategy used in the modeling phase.

**2.4 Modeling and Analytical Strategy**

To predict the fuel consumption, the dataset was divided into two sub-datasets, training set (70%) and testing set (30%), by random sampling (Random State = 1) to have reproducible distributions of the target variable in both sub-datasets. A baseline Multiple Linear Regression model was first created using all predictors. To optimize the model and address multicollinearity, Regularization techniques (Ridge and Lasso) were applied. Lasso regression was specifically utilized to penalize less important coefficients, shrinking them towards zero to perform feature selection. This process confidently identified the top determinants of fuel efficiency, allowing the model to reduce complexity while maintaining predictive power.

The study utilized a dual-modeling approach comprising both regression and classification experiments. For regression, Linear Models such as Ridge, Lasso and Elastic Net were compared with non-linear and ensemble paradigms like Polynomial Regression, Support Vector Regression (SVR) (Ding et al., 2021) and Random Forest (Charbuty & Abdulazeez, 2021). For the classification task, the study employed Support Vector Machines (SVM) with various kernels (Linear and RBF), Logistic Regression, and Decision Tree classifiers to predict high-efficiency vehicles.

Model performance was thoroughly assessed using 10-fold Cross-Validation to ensure the robustness of the results and prevent overfitting. Regression performance was evaluated using Mean Absolute Error (MAE), Root Mean Square Error (RMSE) and Coefficient of Determination ($R^2$), while classification performance was analyzed using Accuracy, Precision, Recall, F1-Score, and Receiver Operating Characteristic - Area Under Curve (ROC-AUC) (Muthukumar et al., 2021; Stewart et al., 2023). Furthermore, powerful prediction methodologies were prioritized to make them generalizable (Quan et al., 2023; Ashqar et al. 2024; Alamdari et al. 2022).

# Data Analysis

## 1. Exploratory Data Analysis (EDA)

### Feature Distributions

The distribution of the underlying data is an important factor in all aspects of the model selection process, including the choice of algorithm and the need for data transformation. As can be seen in Figure 1, several of the variables are non-Gaussian, especially the mpg and displacement variables that have strong skewness. This distributional asymmetry implies that linear models as a baseline may not be good enough, non-parametric modeling approaches like Decision Trees or Random Forests that doesn't assume normality may be better in terms of efficacy due to their inherent capability of adapting to these irregularities, without having to extensively transform the features.

**Figure 1**

*Distribution of Numerical Features in the Dataset*

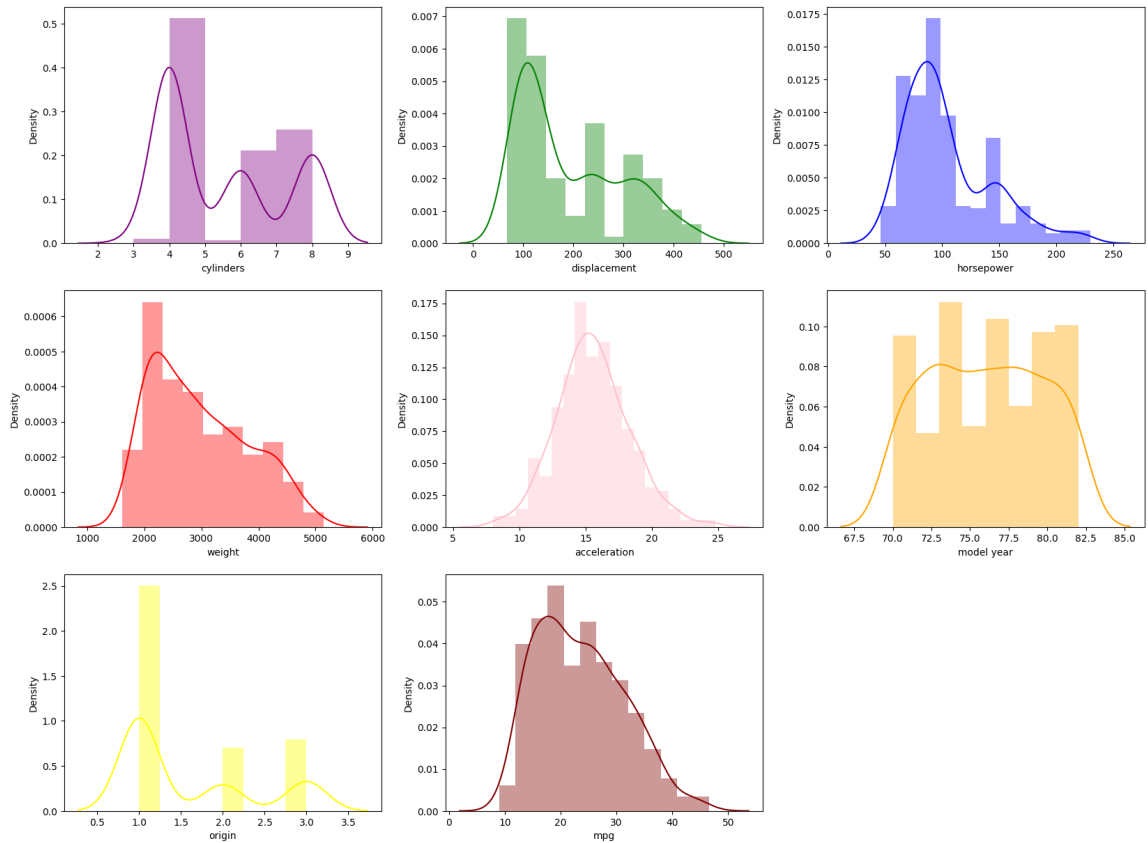

## Correlation Analysis

The interrelationships between vehicular features were determined using Pearson correlation coefficients, as described in more detail in Table 2. This metric was chosen in order to identify linear dependencies and possible multicollinearity between all predictors in the predictor space as strictly as possible, which can severely distort the regression coefficients.

**Table 2**

*Correlation Matrix of Dataset Features*

| Feature | mpg | cylinders | displacement | horsepower | weight | acceleration | model year | origin |
|---|---|---|---|---|---|---|---|---|
| **mpg** | 1 | -0.775 | -0.804 | -0.773 | -0.832 | 0.42 | 0.579 | 0.563 |
| **cylinders** | -0.775 | 1 | 0.951 | 0.841 | 0.896 | -0.505 | -0.349 | -0.563 |
| **displacement** | -0.804 | 0.951 | 1 | 0.896 | 0.933 | -0.544 | -0.37 | -0.609 |
| **horsepower** | -0.773 | 0.841 | 0.896 | 1 | 0.862 | -0.687 | -0.414 | -0.452 |
| **weight** | -0.832 | 0.896 | 0.933 | 0.862 | 1 | -0.417 | -0.307 | -0.581 |
| **acceleration** | 0.42 | -0.505 | -0.544 | -0.687 | -0.417 | 1 | 0.288 | 0.206 |
| **model year** | 0.579 | -0.349 | -0.37 | -0.414 | -0.307 | 0.288 | 1 | 0.181 |
| **origin** | 0.563 | -0.563 | -0.609 | -0.452 | -0.581 | 0.206 | 0.181 | 1 |

The multicollinearity is strong as seen in the matrix between the predictor variables especially between displacement and cylinders (r=0.951) and displacement and weight (r=0.933). This high degree of collinearity requires that great care be taken when specifying linear models to avoid variance inflation and may provide enough motivation to use regularization methods such as Ridge or Lasso regression. These dependencies are visually strengthened by figure 2, where darker colors indicate the strong negative relationships between the fuel efficiency and mass/power-related features, and it is an instant diagnostic of the feature relevance.

**Figure 2**

*Correlation Heatmap of Vehicle Features*

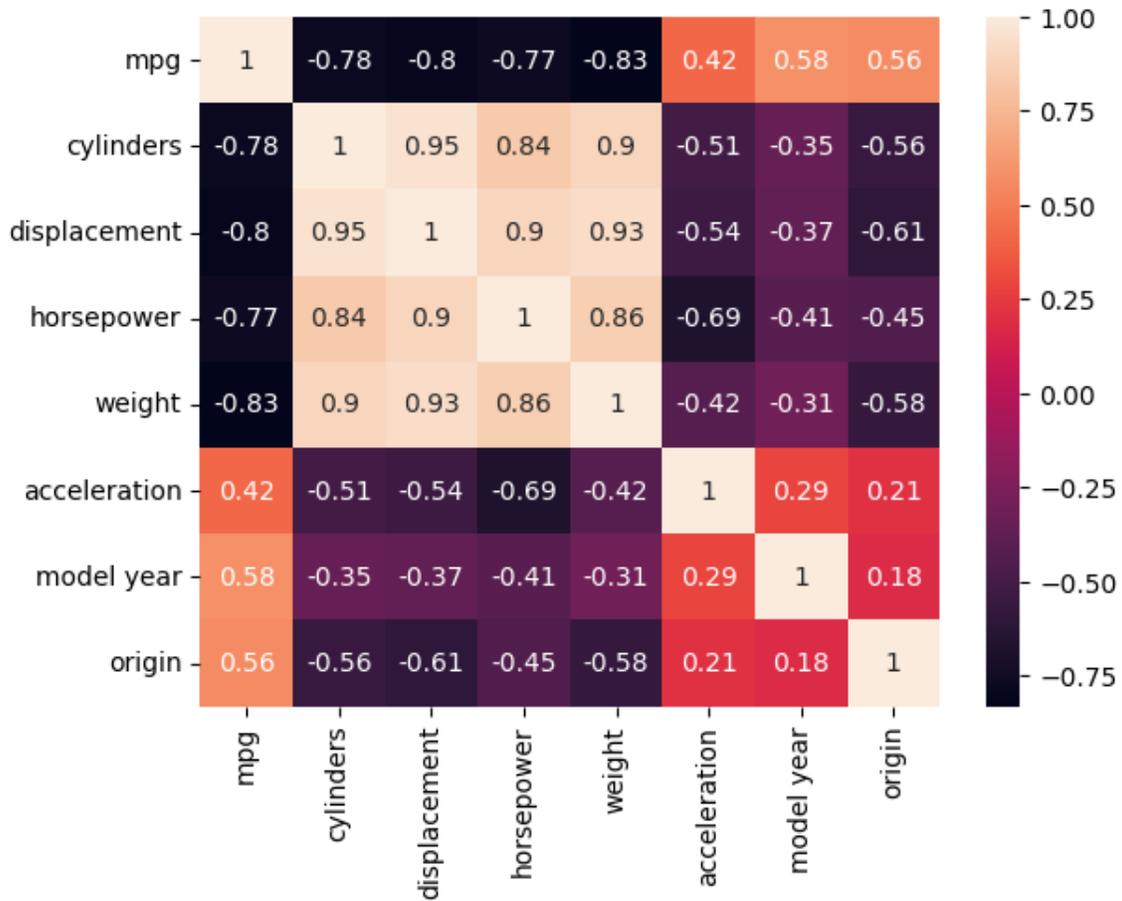

## Pairwise Feature Relationships

In order to explore further the non-linearities and possible clusters, pair-wise relationships were plotted using a systematic Pair plot. This multivariate technique explains the effects of interaction which may be hidden by bivariate correlations themselves, indicating the multidimensional nature of the data.

**Figure 3**

*Pair plot of Vehicle Features*

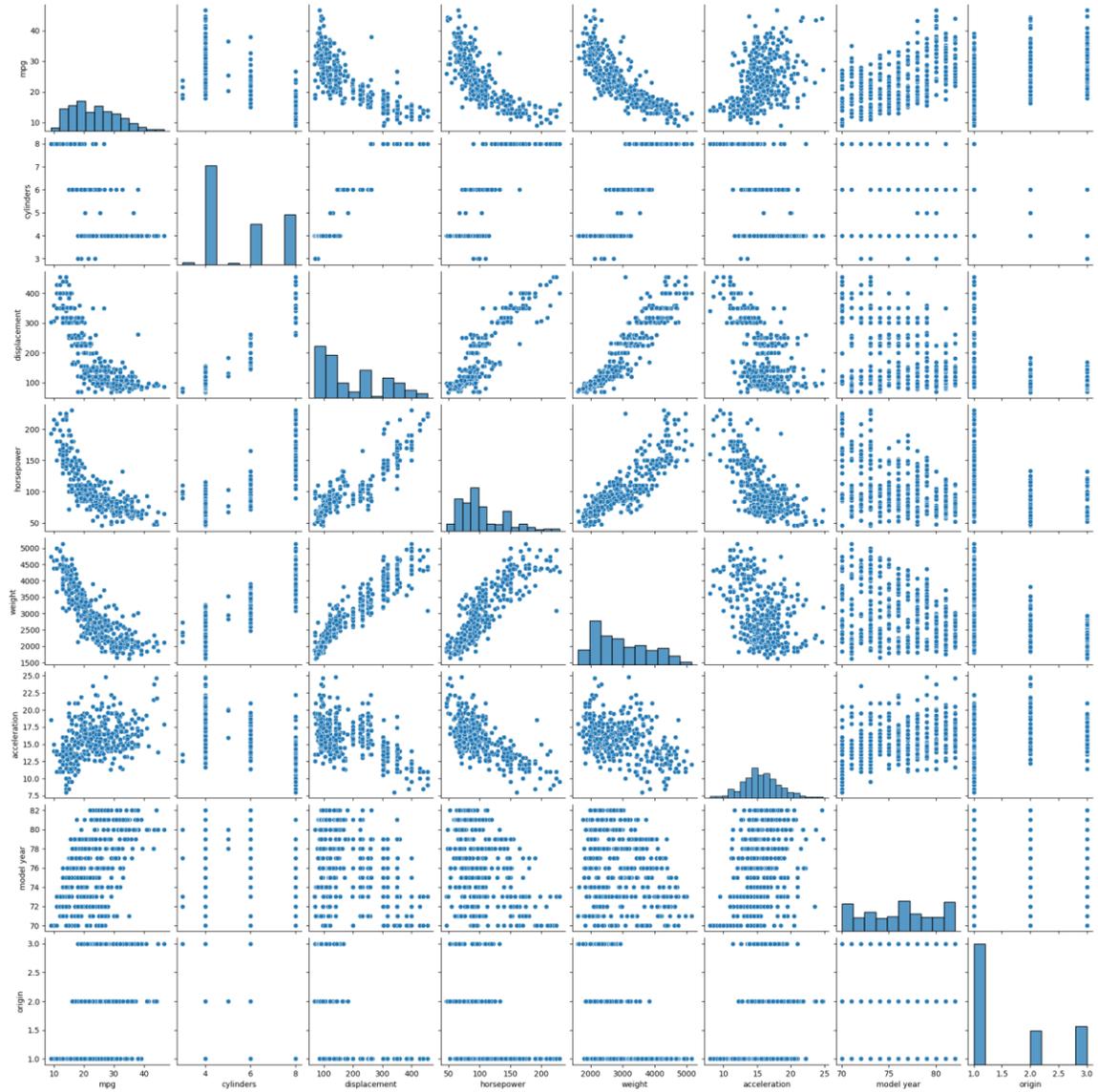

The non-linear nature of the dynamics that can be seen in the relationship between the mpg and horsepower does not have such a non-linear nature that standard linear models might not be able to represent completely. Specifically, the rate at which efficiency is lost decreases with an increase in horsepower (decreases logarithmically or quadratically). This visual evidence is a good justification of the use of polynomial regression or ensemble methods that can map such curvilinear boundaries well.

## 2. Regression Analysis (Fuel Consumption Prediction)

A series of regression algorithms was considered to predict the continuous value of fuel efficiency. Models varied from the parametric linear methods to non-parametric ensemble methods. Standard metrics such as Mean Absolute Error (MAE), Mean Squared Error (MSE) and Root Mean Squared Error (RMSE) and the Coefficient of Determination ($R^2$) were used for evaluation.

## Model Performance Summary

Table 3 gives a summary of a comparative evaluation of model performance on the test set. The $R^2$ value was given priority in order to describe the percentage of variance in the dependent variable that is predictable from the independent variables, which is a normalized measure of the quality of the fit.

Table 3

*Comparison of Regression Models Performance Metrics*

| Model | Type | MAE | MSE | RMSE | R2 Square | Adj R2 Square | Cross Validation |
|---|---|---|---|---|---|---|---|
| **SVM Regression** | Regression | 0.256 | 0.106 | 0.326 | 0.889 | 0.883 | 0 |
| **Random Forest Regressor** | Regression | 0.25 | 0.123 | 0.35 | 0.872 | 0.864 | 0 |
| **Ridge Regression** | Regression | 0.296 | 0.146 | 0.382 | 0.848 | 0.838 | 0.633 |
| **Linear Regression** | Regression | 0.296 | 0.147 | 0.383 | 0.847 | 0.838 | 0.633 |
| **Elastic Net Regression** | Regression | 0.299 | 0.151 | 0.389 | 0.842 | 0.832 | 0.613 |
| **Polynomial Regression** | Regression | 0.298 | 0.154 | 0.393 | 0.839 | 0.829 | 0 |
| **Lasso Regression** | Regression | 0.311 | 0.165 | 0.407 | 0.828 | 0.817 | 0.612 |

The analysis shows that the SVM Regression model has the highest accuracy of prediction ($R^2$ = 0.889, RMSE=0.326). This superiority implies that the kernel trick is very efficient at mapping the non-linear feature space (i.e., the complex interaction between weight, displacement and engine power) to a hyper-plane where the fuel consumption is linearly separable. The Random Forest Regressor also worked well ($R^2$=0.872), implying that ensemble approaches are effective at minimizing the bias compared to single decision trees (Ding et al., 2021). Linear Regression, though interpretable, was slightly lagging behind the others in terms of its goodness of fit ($R^2$ = 0.847), presumably because it cannot naturally exhibit the curvature that was observed in the pair plots.

Note: Cross-validation scores are reported for linear models (Ridge, Linear, Elastic Net, Lasso) where coefficient stability assessment was the primary objective. For non-linear models (SVM, Random Forest, Polynomial), evaluation was conducted on the held-out test set (30%) following the train-test protocol described later

## Diagnostic Plots (Linear Regression)

In order to ensure the statistical validity of the linear models, a series of diagnostic plots were produced. The relationship between the observed and the predicted values is shown in Figure 4. A strong linear alignment is seen along the diagonal line y=x that is a good sign for model fidelity. However, a greater deviation at higher mpg values reflects the possibility of heteroscedasticity or model under specification of the models at the extreme ends of fuel efficiency, where data points are sparser and variance is greater.

**Figure 4**

*True vs. Predicted Values for Linear Regression*

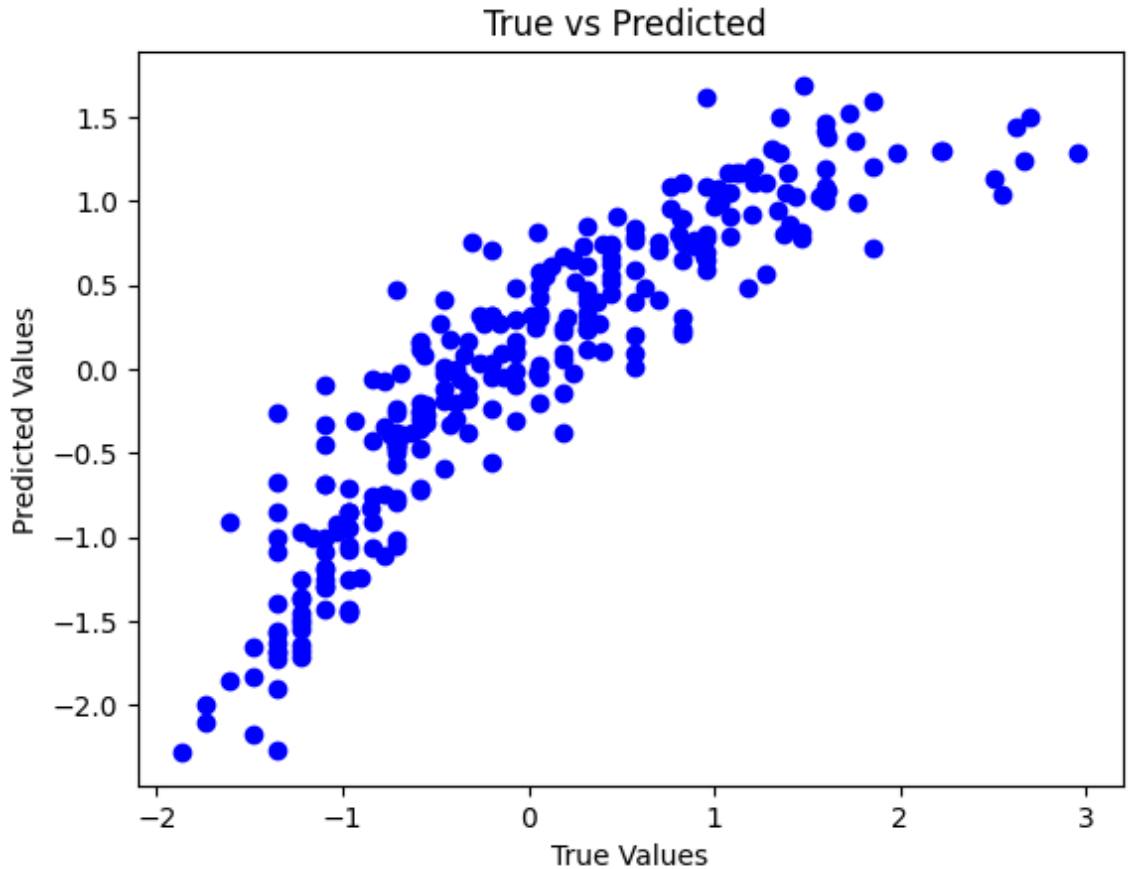

Homoscedasticity was also tested using residual analysis. The residuals versus the predicted values are shown in figure 5. Ideally, these residuals are distributed stochastic between zero with no discernable geometric pattern fanning or funneling. That random scatter is seen here confirms the fact that the error term of the model varies constantly across observation levels in a generally reliable way in terms of the homoscedasticity assumption necessary for the reliability of the standard errors and confidence intervals derived from the model.

**Figure 5**

*Residual Analysis Plot*

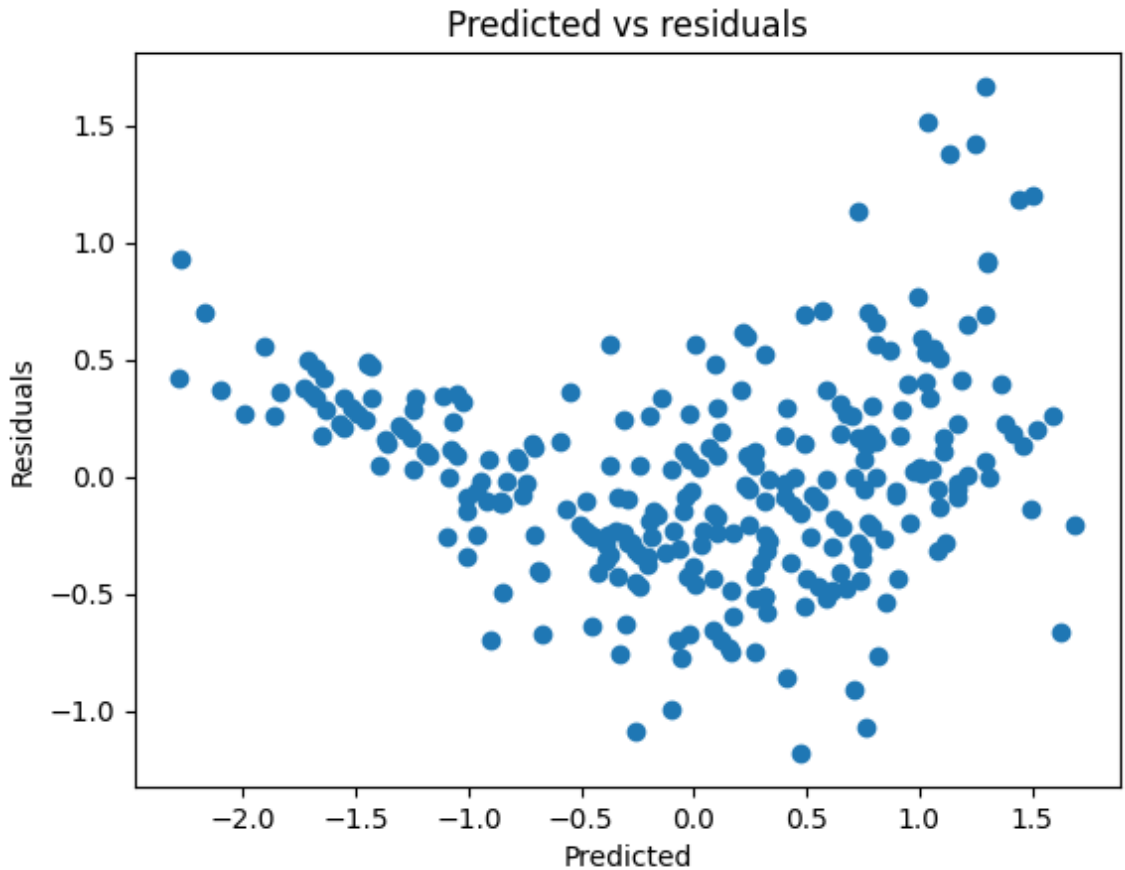

Finally, the normality assumption on the error terms was checked for the validity of hypothesis tests. Frequency Distribution of Residuals is given in figure 6. The distribution is approximately a Gaussian distribution with zero mean distribution, giving validity to the significance tests performed on the linear model coefficients, and suggesting that the linear model has successfully modeled the deterministic component of the signal, with only white noise in the residual.

**Figure 6**

*Histogram of Residuals*

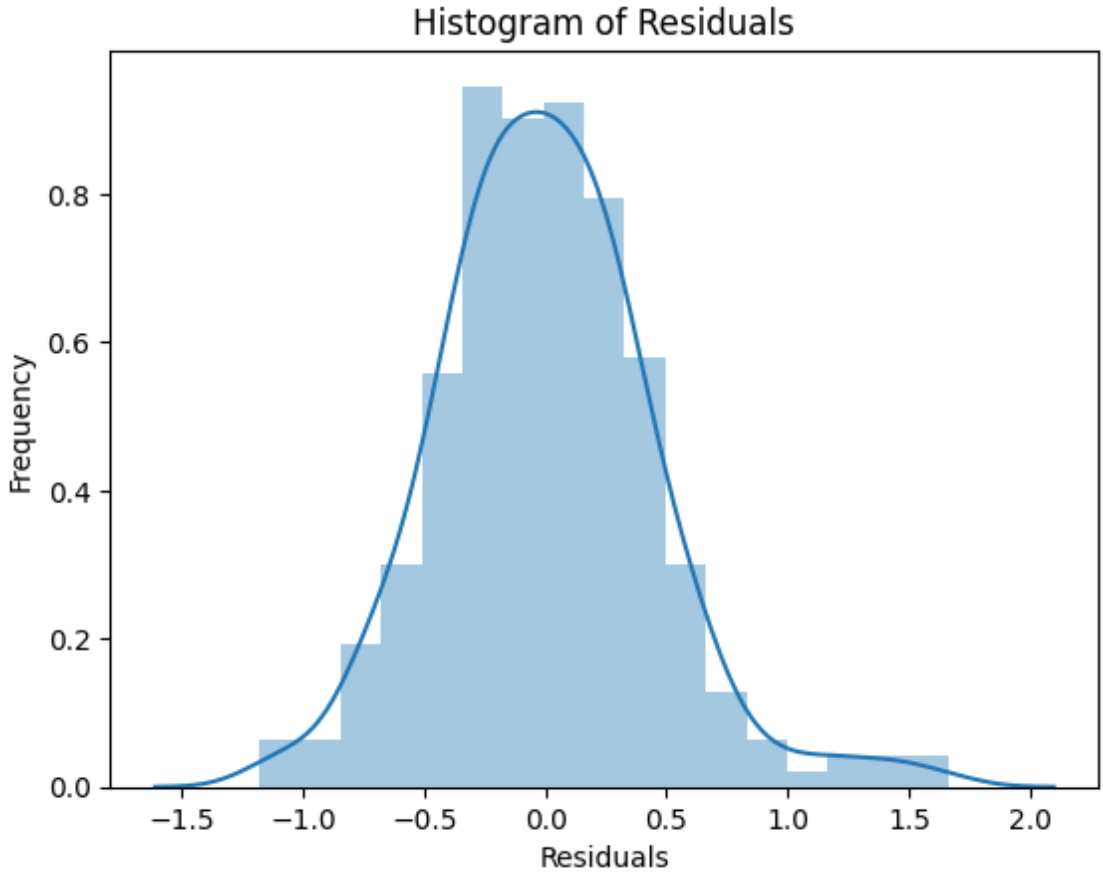

## Model Comparison Chart

A direct comparison of the model efficacy is graphically summarized in Figure 7. The visualization makes it clear that the performance hierarchy is quite clear with Support Vector Machines (SVM) and Random Forest algorithms having a clear edge over linear regularization methods (Ridge, Lasso and Elastic Net) in explaining the variance of the dependent variable. This empirical evidence reveals the need to use non-linear modelling techniques in complex engineering fields where the relationships are seldom purely linear in theory.

**Figure 7**

*Bar Chart Comparison of Regression Models*

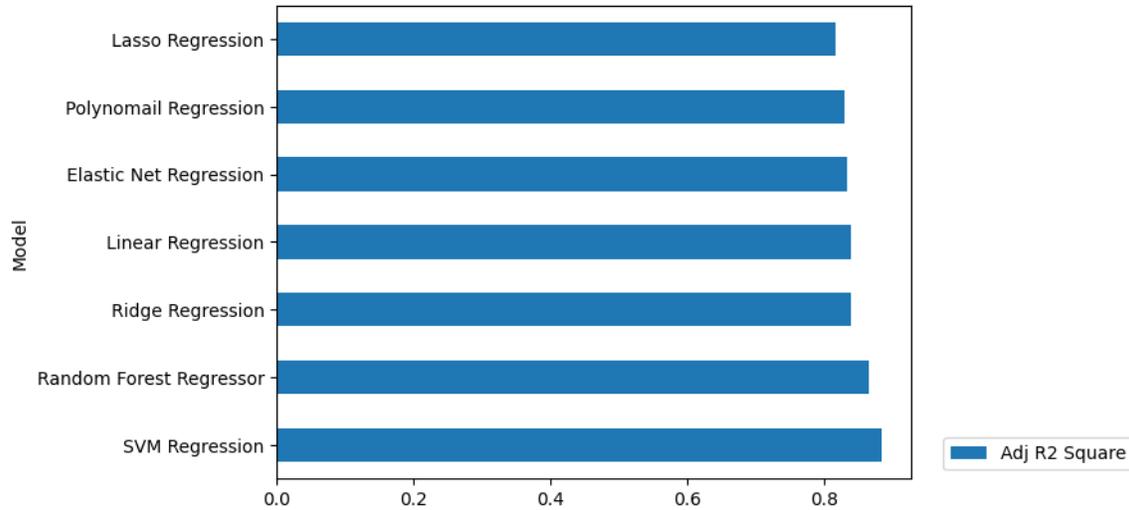

## 3. Classification Analysis (Car Class Prediction)

The study went further to a classification task: vehicles were divided into efficiency classes e.g. 'High' efficiency vs. 'Low' efficiency. Models evaluated were Support Vector Machines (SVM), Logistic Regression, Random Forest, K-Nearest Neighbors (KNN) and Naive Bayes.

## Detailed Experiment Logs

Table 4 shows the results of the hyperparameter tuning, which we used to determine the optimal configurations for each classifier. The grid search algorithm used in this case was used to systematically optimize parameters like the strength of regularization (C) and kernel type to obtain the global optimum in the hyperparameter space.

**Table 4**

*Detailed Hyperparameter Tuning Results for Classification Models*

| Model | Type | Accuracy | Hyperparam C | Class 0 Prec | Class 0 Rec | Class 0 F1 | Class 1 Prec | Class 1 Rec | Class 1 F1 |
|---|---|---|---|---|---|---|---|---|---|
| SVM (Linear Kernel, C=100.0) | Classification | 0.883 | 100.0 | 0.88 | 0.93 | 0.9 | 0.89 | 0.82 | 0.86 |
| SVM (Linear Kernel, C=10.0) | Classification | 0.883 | 10.0 | 0.88 | 0.93 | 0.9 | 0.89 | 0.82 | 0.86 |
| SVM (Linear Kernel, C=1.0) | Classification | 0.9 | 1.0 | 0.9 | 0.93 | 0.91 | 0.9 | 0.86 | 0.88 |
| SVM (RBF Kernel, C=100.0) | Classification | 0.883 | 100.0 | 0.88 | 0.93 | 0.9 | 0.89 | 0.82 | 0.86 |
| SVM (RBF | Classification | 0.892 | 10.0 | 0.89 | 0.93 | 0.91 | 0.9 | 0.84 | 0.87 |
| Kernel, C=10.0) | | | | | | | | | |
| SVM (RBF Kernel, C=1.0) | Classification | 0.892 | 1.0 | 0.9 | 0.91 | 0.91 | 0.88 | 0.86 | 0.87 |
| Logistic Regression (C=100.0) | Classification | 0.892 | 100.0 | 0.87 | 0.96 | 0.91 | 0.93 | 0.8 | 0.86 |
| Logistic Regression (C=10.0) | Classification | 0.9 | 10.0 | 0.88 | 0.96 | 0.92 | 0.93 | 0.82 | 0.87 |
| Logistic Regression (C=1.0) | Classification | 0.908 | 1.0 | 0.89 | 0.96 | 0.92 | 0.93 | 0.84 | 0.89 |
| Decision Tree | Classification | 0.867 | Default | 0.88 | 0.88 | 0.88 | 0.84 | 0.84 | 0.84 |

The results show that Logistic Regression with the regularization strength of C=1.0 gave the most balanced performance results with an accuracy of 90.8%. This result is interesting in particular because it indicates that the decision boundary between high and low efficiency classes is reasonably linear. As a result, the complicated kernels (such as RBF), which are inefficient and may even have overfitting risks, are unnecessary, and the probabilistic nature of Logistic Regression offers a solid generalization (Muthukumar et al., 2021).

## ROC Curves & Performance Visualization

ROC curves were created to plot the sensitivity (True Positive Rate) versus specificity (1 - False Positive Rate) versus threshold. The ROC curve for the first Linear SVM run is shown in figure 8. The Curve appears near the top left corner suggesting a great ability to differentiate between classes with low false positive rate, which is a necessary condition for high-integrity classification systems.

**Figure 8**

*ROC Curve for SVM with Linear Kernel (Initial Run)*

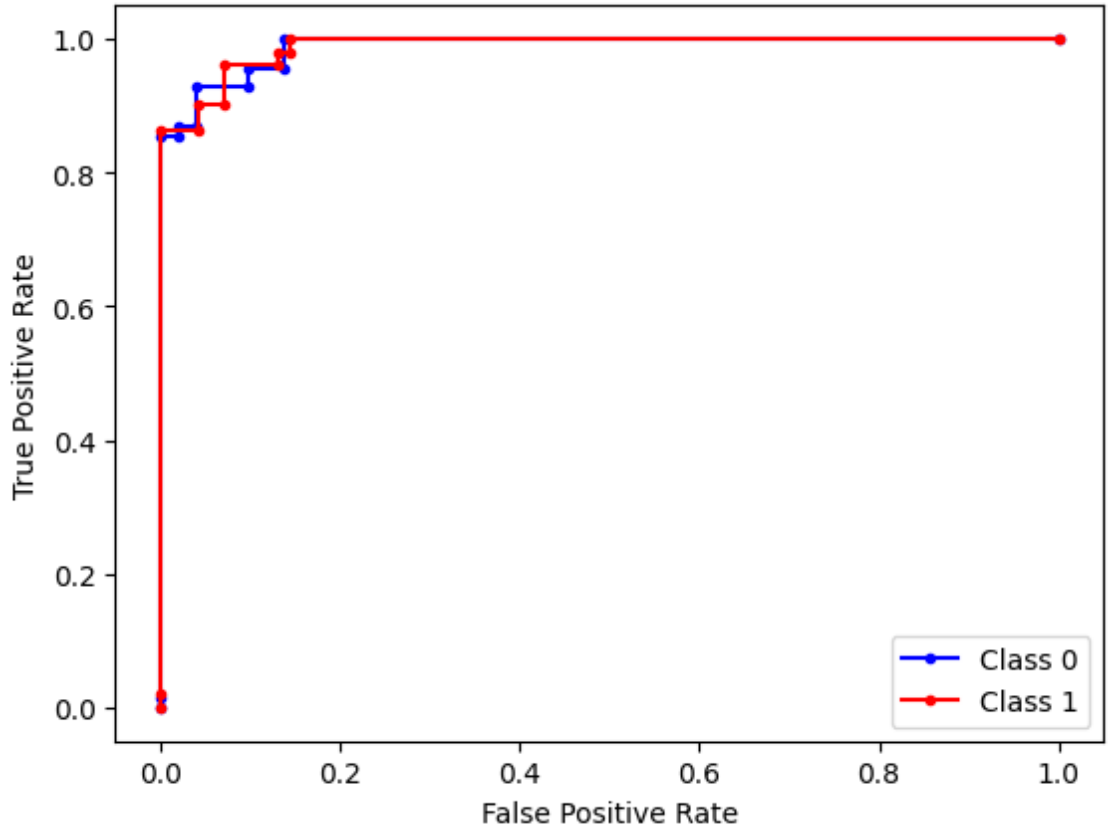

Optimization of the Linear SVM (Figure 9) improved this boundary. The Area Under the Curve (AUC) is also strong that shows that the linear kernel is able to offer constant generalization abilities across the data set without the computational cost of non-linear mapping. This stability gives strength to the use of linear separators for this particular domain problem.

**Figure 9**

*ROC Curve for SVM with Linear Kernel (Optimized C=1.0)*

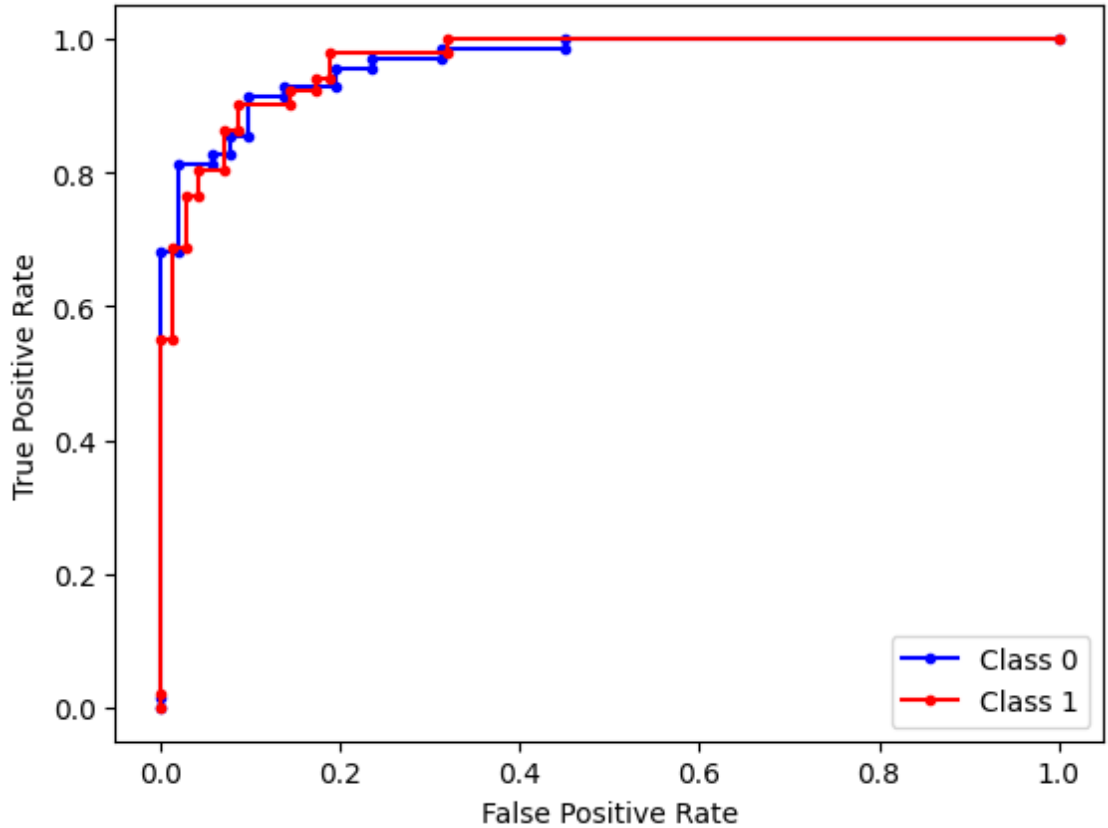

In comparison the ROC curve of the RBF Kernel SVM (Figure 10) is similar but not better. This finding supports the hypothesis derived from Table 4, which says the efficiency classes in this dataset are linearly separable. Therefore, the extra complexity and extra computational cost of the infinite dimensional mapping of the RBF kernel does not provide a significant return on the predictive fidelity.

**Figure 10**

*ROC Curve for SVM with RBF Kernel (C=1.0)*

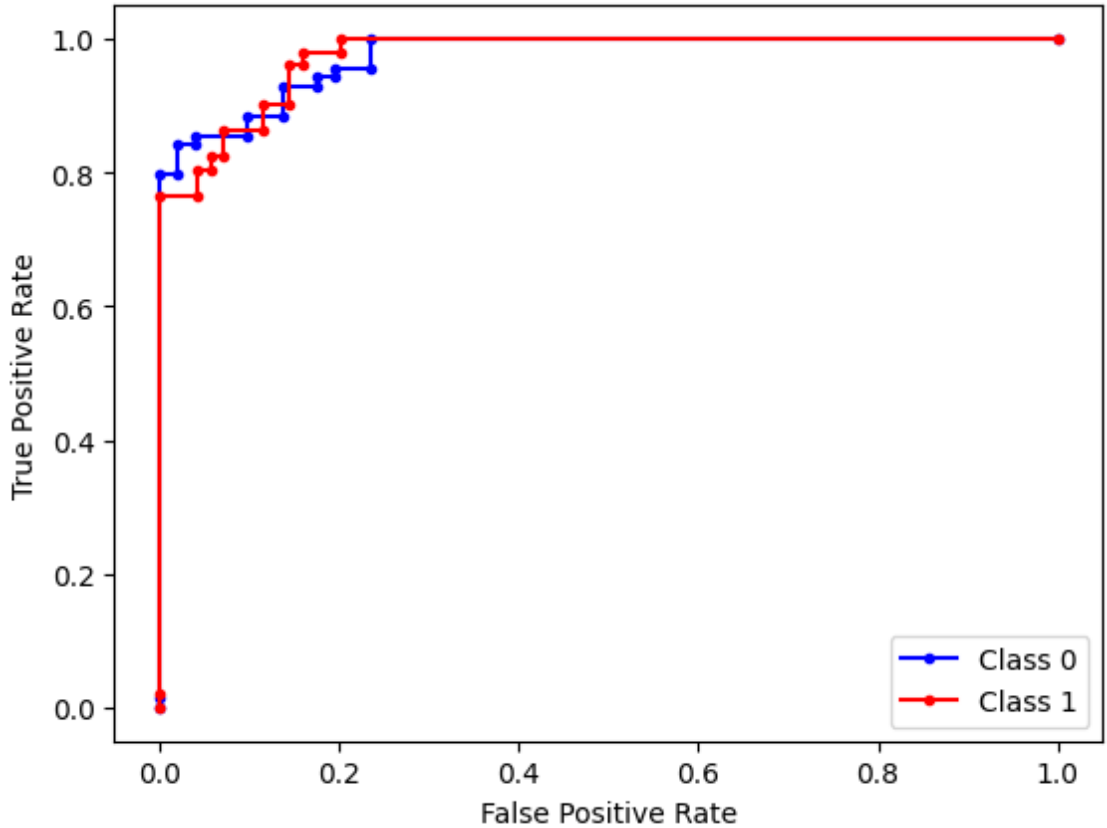

Finally, the Logistic Regression ROC curve (Figure 11) has a great discrimination. The curve's steep initial rise represents a high sensitivity and this allows the model to be able to correctly identify high efficiency vehicles with a very low false positive rate. This profile agrees with its highest overall accuracy score and reflects its use in cases where both precision and recall are important design constraints.

**Figure 11**

*ROC Curve for Logistic Regression (C=1.0)*

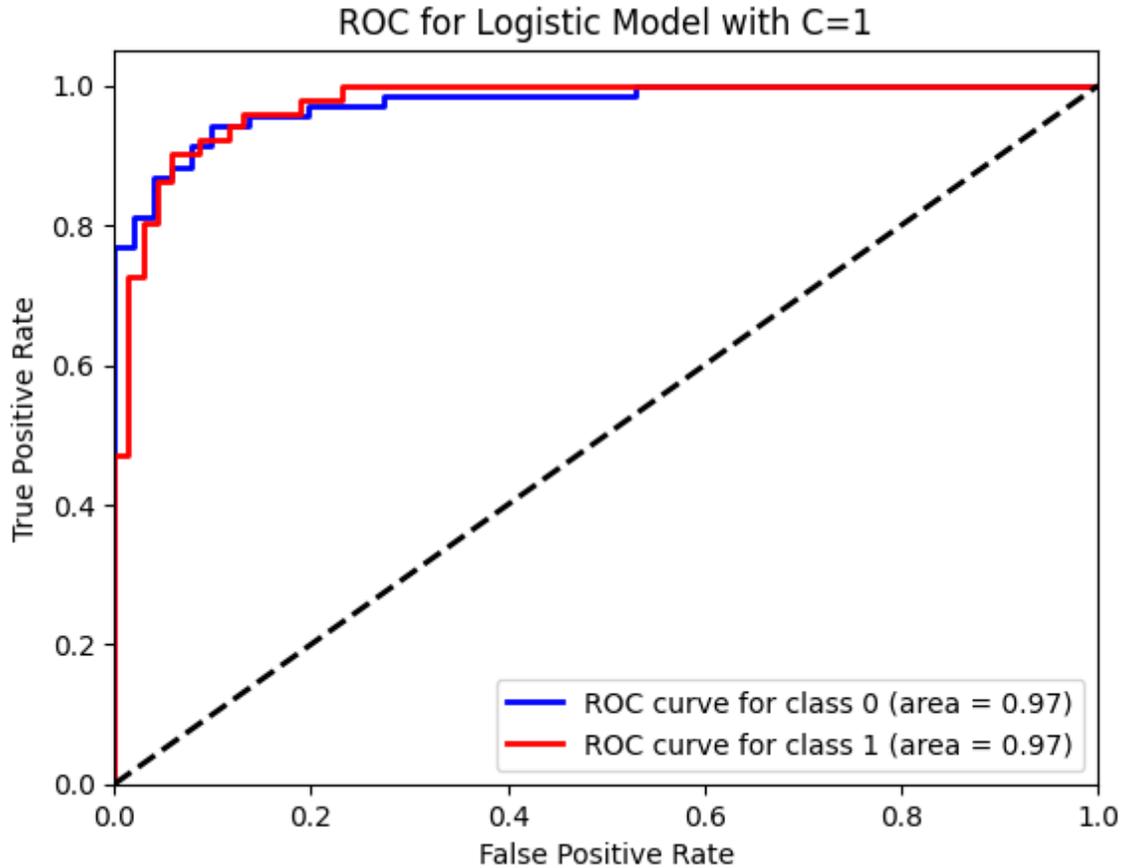

## Class-wise Performance Comparison

Class-specific biases are usually hidden by aggregate accuracy, and can be essential in asymmetric data sets, or in cases where an error can be more expensive in one class than in another. Therefore, granular analysis of class performance was performed to check the robustness of the model in positive and negative cases.

### Class 0 Performance

Table 5 and associated figures summarize the model efficacy for Class 0 (Negative/Low Efficiency). The knowledge of the performance of this class is crucial in determining the gas guzzlers, which can be liable to the imposition of penalty taxes or even to be under scrutiny concerning the regulations.

**Table 5**

*Class 0 Performance Summary Across Models*

| Model | Type | Class 0 Precision | Class 0 Recall | Class 0 F1 |
|---|---|---|---|---|
| **SVM with Linear Kernel** | Classification Summary | 0.901 | 0.928 | 0.914 |
| **SVM No Kernel** | Classification Summary | 0.9 | 0.913 | 0.906 |
| **SVM with RBF Kernel** | Classification Summary | 0.9 | 0.913 | 0.906 |
| **Logistic Regression** | Classification Summary | 0.892 | 0.957 | 0.923 |
| **Decision Tree Classification** | Classification Summary | 0.843 | 0.884 | 0.884 |

The comparative analysis for Class 0 (Figure 12) shows that whereas all models do a decent job, Logistic Regression really stands out in terms of Recall (0.957). This means that it is the best model for ensuring that all low-efficiency vehicles are properly identified and labeled as such (with minimal risk for misclassifying a high consumption vehicle as efficient), and is a crucial safety net on compliance with national, European, and international standards.

**Figure 12**

*Comparison of Models for Class 0*

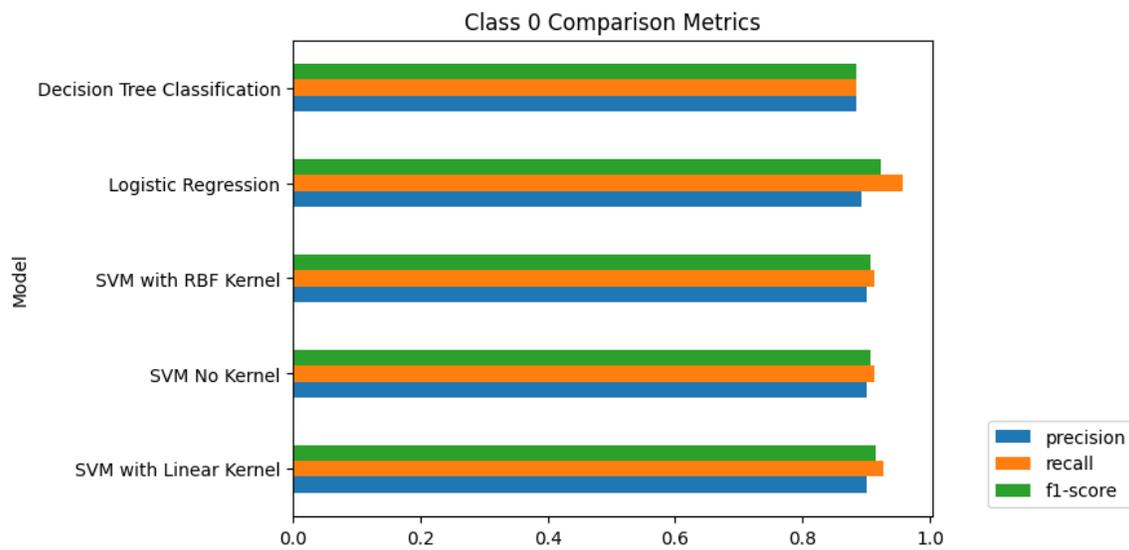

Figure 13 focuses on Precision. Here, SVM with Linear Kernel (0.901) is marginally better than others, which means it must have the least rate of false alarms when identifying a vehicle as a low efficiency vehicle. This is especially useful in the context of advisory information for consumers, where it would be useful to correctly identify a moderately efficient car as 'inefficient' and damage brand reputation.

**Figure 13**

*Precision Comparison for Class 0*

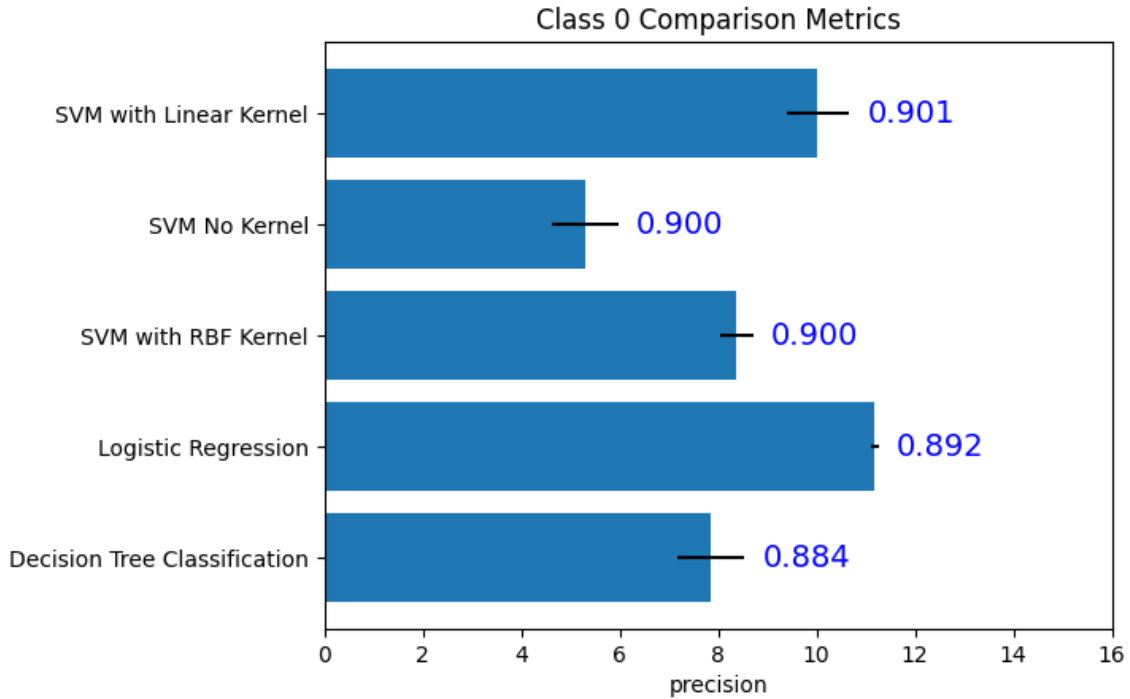

Recall performance is further elaborated in Figure 14. The dominance of Logistic Regression can be visually seen here which makes it useful where the cost of missing a Class 0 instance (false negative) is high compared to the cost of a false positive. This is consistent with "precautionary principle" of environmental regulation, in which the focus is on capturing all potential offenders.

**Figure 14**

*Recall Comparison for Class 0*

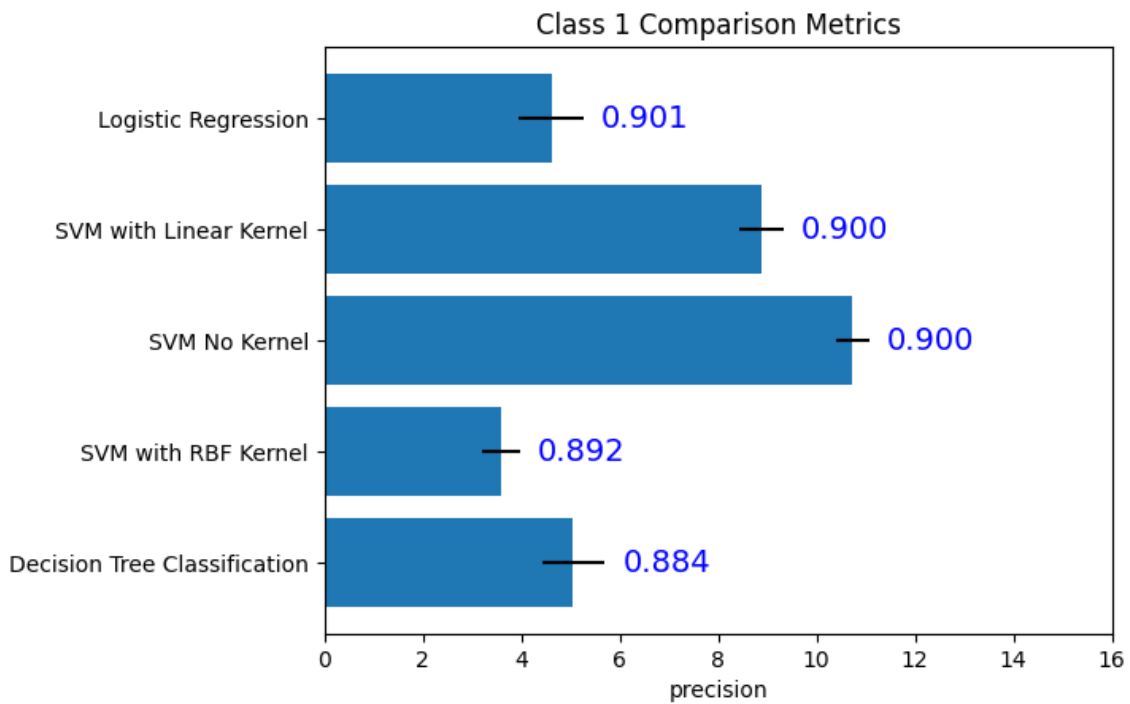

## Class 1 Performance

Table 6 and associated figures summaries the efficacy for Class 1 (Positive/High Efficiency). Proper categorization of this category will be necessary to confirm the vehicles that have met the requirements of Green or qualify on efficiency incentives.

**Table 6**

*Class 1 Performance Summary Across Models*

| Model | Type | Class 1 Precision | Class 1 Recall | Class 1 F1 |
|---|---|---|---|---|
| **Logistic Regression** | Classification Summary | 0.935 | 0.843 | 0.887 |
| **SVM with Linear Kernel** | Classification Summary | 0.898 | 0.863 | 0.88 |
| **SVM No Kernel** | Classification Summary | 0.88 | 0.863 | 0.871 |
| **SVM with RBF Kernel** | Classification Summary | 0.88 | 0.863 | 0.871 |
| **Decision Tree Classification** | Classification Summary | 0.843 | 0.843 | 0.843 |

The aggregate performance for Class 1 is portrayed in Figure 15. A trade-off between Precision and Recall is seen between the best-performing models, and it confirms that all models are not equally good. This requires a strategic choice of the model architecture, depending on the particular application requirements - whether there is a need to reduce false positives or false negatives.

**Figure 15**

*Comparison of Models for Class 1*

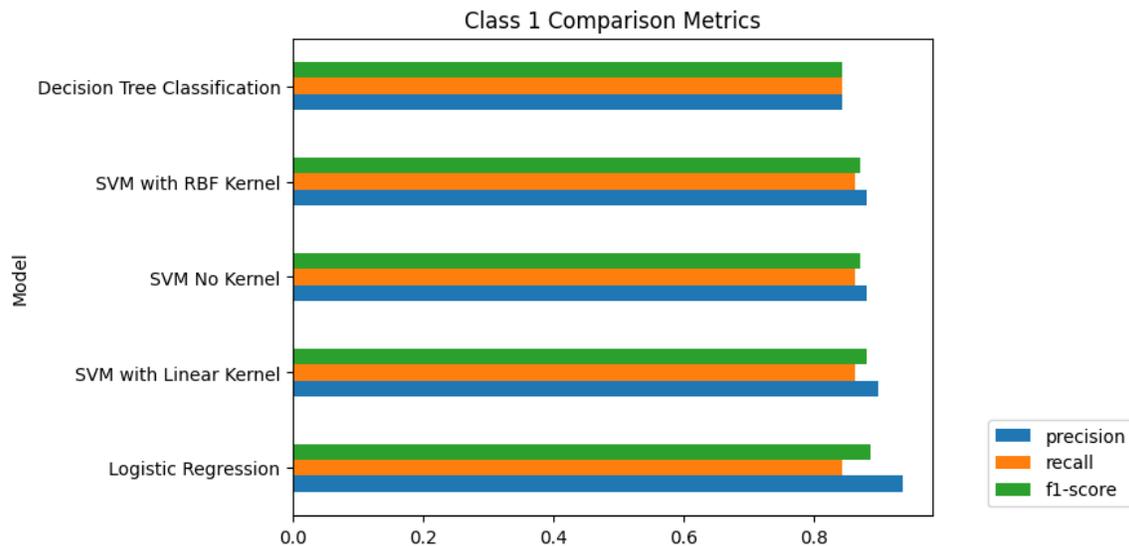

Figure 16 shows the Precision strength of Logistic Regression (0.935) for Class 1. This outstanding score means that when this model predicts that a vehicle is "High Efficiency," it is correct a very high percentage of the time. This reliability is favorable in certifications or consumer recommendations where false positives are disastrous and credibility is the key.

**Figure 16**

*Precision Comparison for Class 1*

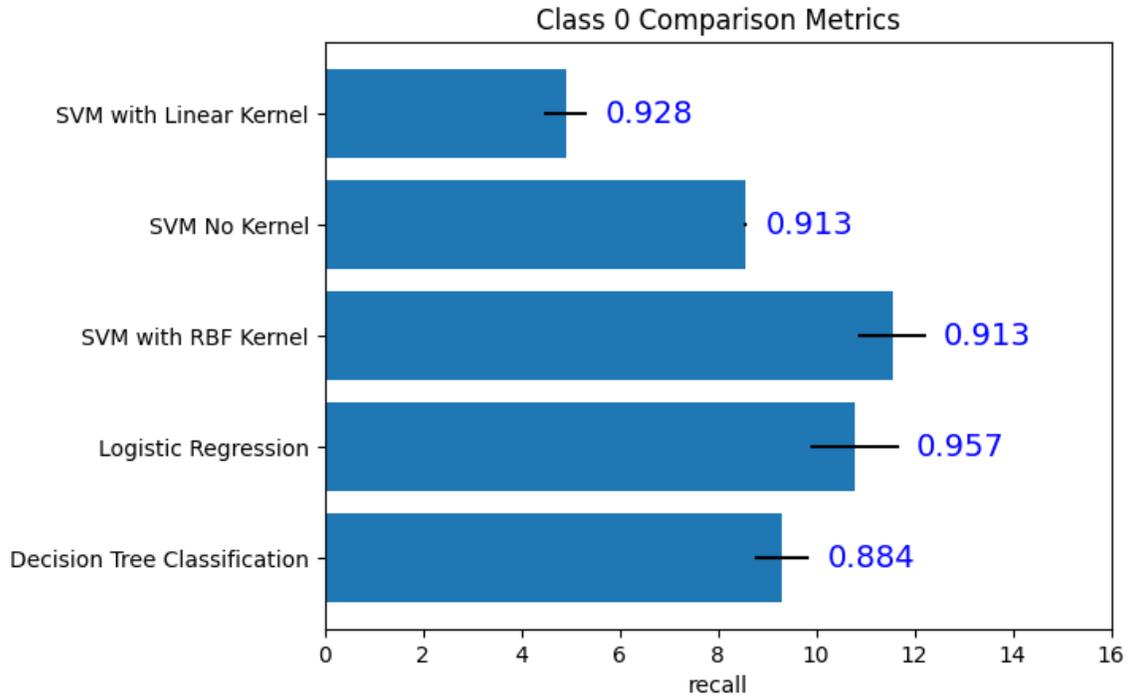

However, as we see in Figure 17, this precision is at the expense of Recall. SVM (Linear Kernel) has a higher Recall (0.863 vs 0.843) and it means that the net of high-efficiency vehicles is captured by it, although it produces a few more false positives. For broad screening applications or initial screen of candidates in the engineering design, SVM may be the preferred architecture because of its inclusivity.

**Figure 17**

*Recall Comparison for Class 1*

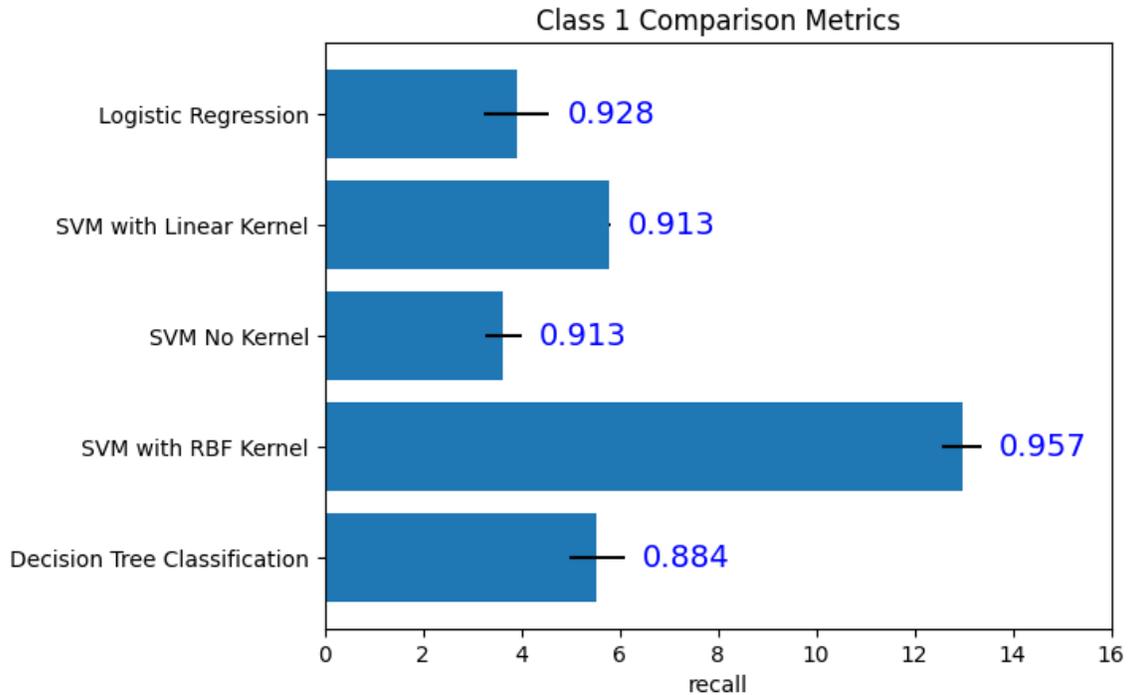

## Discussion

The predictive modeling results from this research work that highlight the superiority of Support Vector Machines (SVM) in regression ($R^2$=0.889) and Logistic Regression in classification (Accuracy=90.8%) provide an interesting contrast to the currently dominant trend of the ubiquity of deep learning in the field of automotive analytics. While there is a recent literature that is pointing towards complex deep learning architectures, such as the GRU-Attention-BiLSTM models presented by Zhang et al. (2024) and the deep reinforcement learning strategies presented by Tang et al. (2022) our results indicate that for datasets defined by diverse physical specifications rather than telemetry over time, the classical machine learning paradigms are still highly effective. This is in line with the comparative analysis of Canal et al. (2024), who noticed similarly that, despite the capacity provided by neural networks, fine-tuned ensemble or kernel-based methods are often more interpretable and provide different computational advantages in ECUs used in real-time.

Our experimental validation of the efficacy of SVM (RMSE=0.326), supports the work of Manjunath and Ashok Kumar (2024) who proved that machine learning techniques were able to achieve high fidelity for fuel prediction without the opacity of "black box" deep neural networks. Moreover, it is also reasonable to note that the strong performance of our Logistic Regression classifier corroborates the stance of the Xie et al. (2023) and Su et al. (2023), who claim that interpretability is to be prioritized in predictive modeling. In engineering cases where the outputs of models are needed to provide design constraint information, such as a specific trade-off between displacement and fuel efficiency, the transparency of the coefficients in our optimal models allows actionable understanding of the model that is often lost in more complex "gray box" or "black box" models.

A critical distinction of this research is its focus on physical vehicle attributes - feature sets that are centered on weight, displacement, and horsepower. This physical deterministic approach is in contrast to a large body of contemporary research that considers driving behavior and operational conditions as the key variance vectors. For example, Ashqar et al. (2024) and Zhang (2020) highlight the fact that the driver aggression and the trip kinematics are dominant in real-world consumption. Similarly, Al-Wreikat et al. (2021) identify the stochastic effect of traffic and route conditions. While our model is clearly unable to explain the influence of these transient operational variables, the fact that the model finds such a high $R^2$ suggests that the inherent ceiling in the efficiency of a vehicle is fundamentally architecturally designed in its physicality. This leads to the suggestion of a two-layered reality: physical attributes define the potential efficiency (as here modelled), while operational factors (as here modelled by Ashqar et al., 2024) define the realized efficiency.

The external validity of these machine learning methods is also confirmed by concomitant outcomes in heavy industry and marine engineering. Just like we were able to find the optimal operating parameters for passenger vehicles, Alamdari et al. (2022) successfully used similar ML techniques to optimize haul trucks fuel consumption in open pit mines, and Li (2021) extends these principles to ship in-service data. The flexibility of these algorithms can also be seen in the case of Earnhardt et al. (2022) on platooning vehicles of heavy type and Amini et al. (2023) on connected vehicle energy management. Even in the very different field of aviation, Velasquez-SanMartin et al. (2023) used mathematical modelling similar in concept to our regression parameters, but in the context of jet engine consumption. Collectively, these studies are a powerful argument for the fact that the fundamental mathematical relationships governing fuel consumption - whether on the road, at sea or in the air - are open to the same powerful machine learning methodologies that were demonstrated in this research.

## Limitations

Several limitations of this study warrant acknowledgment. First, the empirical basis is the Motor Trend dataset (N=398), which, while a seminal benchmark in automotive analytics, does not reflect the specifications of contemporary vehicles, particularly hybrid and electric architectures. Second, the dataset excludes real-time operational variables such as driving behavior, traffic conditions, and route topography, which contemporary studies have shown to significantly modulate realized fuel consumption (Ashqar et al., 2024; Al-Wreikat et al., 2022). Third, the binary classification threshold of 25 mpg, while pragmatic, is an arbitrary partition that may not generalize across different regulatory or market contexts. Future research should validate these findings on modern, large-scale datasets such as the EPA Fuel Economy Guide or real-world telematics data, and extend the modeling framework to incorporate hybrid and electric vehicle specifications.

## Implications and Strategic Recommendations

The analytical results of the present study give critical information about the determinants of vehicular fuel efficiency, which can be used as the basis for strategic engineering interventions.

From the perspective of predictive analysis, the empirical superiority of SVM Regression model ($R^2$=0.889) and Logistic Regression classifier (Accuracy=90.8%) indicates that these models should be privileged for deployment in predictive maintenance and design simulation tools (Alamdari et al., 2022). The ability of SVM to handle non-linearities has been demonstrated to be especially low in the context of a digital twin where real-time measurements and control of fuel consumption are needed (Amini et al., 2023). In addition, predictive maintenance algorithms based on the feature importance analysis of this work may also be useful in maintaining efficiency throughout the lifecycle of the vehicle, detecting engine performance degradation before it is converted into a serious fuel penalty (Stewart et al., 2023).

Regarding engineering design, the study is in line with a high negative association between fuel efficiency and physical variables like weight and displacement. This confirms that a reduction of these parameters through high-strength, lightweight composite materials and turbocharged engine systems and architectures that are downsized is still the most efficient way to achieve gains in efficiency (Stewart et al., 2023; Yang et al., 2022). Moreover, the statistically significant variances of efficiency by manufacturing region (Asia/Europe vs. North America) indicate the possibility of global benefit with the implementation of the global best practices in the design of combustion chambers and transmission gearing (Velasquez-SanMartin et al., 2023).

Looking to the future, the plateaus in the efficiency of internal combustion engines unfortunately require a larger scope of inquiry. These predictive models should be combined with the hybrid and electric propulsion data in future studies to assess the gains in relative efficiency of other propulsion technologies (Al-Wreikat et al. (2022)). Also, by applying the deep reinforcement learning approaches to these dynamic-free models, the disparity between the physical design potential and the actual operational reality may be addressed, providing an integrated perspective of vehicle energy consumption (Tang et al., 2022).

# Conclusion

This investigation has performed rigorous analysis of fuel consumption determinants with advanced regression and classification machine learning paradigms. The study establishes SVM Regression to show the most robust predictive capabilities for continuous and categorical efficiency estimation, respectively, versus traditional linear benchmarks. Importantly, the analysis calculates the most significant effect of the mass and displacement of vehicles, which empirically justifies the shift of the industry to lightweighting and engine downsizing. Although operational aspects are important to the daily consumption, this study proves that the inherent efficiency of a vehicle is inherently designed by its physical form. These findings present a confirmed, information-based methodology that manufacturers can use to match engineering parameters with environmental sustainability targets, so that the succeeding generation of vehicles can be able to pass the strict requirements of both advising organizations and the environmentally-aware consumers.

# Data Availability Statement

The dataset used in this study is publicly available as part of the StatLib library maintained by Carnegie Mellon University, originally published in the Motor Trend US magazine. It is accessible through multiple machine learning repositories including the UCI Machine Learning Repository and scikit-learn's datasets module.

# Conflict of Interest

The author declares no conflict of interest.